\def\year{2021}\relax
\documentclass[letterpaper]{article} 
\usepackage{aaai21}  
\usepackage{times}  
\usepackage{helvet} 
\usepackage{courier}  
\usepackage[hyphens]{url}  
\usepackage{graphicx} 
\usepackage{natbib}  
\usepackage{caption} 
\usepackage{algorithm}
\usepackage{algorithmic}
\usepackage{mathrsfs}
\usepackage{amsmath}
\usepackage{graphicx}
\usepackage{booktabs}
\usepackage{array}
\usepackage{boldline}
\newcolumntype{C}[1]{>{\centering}m{#1}}

\title{A Student-Teacher Architecture for Dialog Domain Adaptation under the Meta-Learning Setting}
\author {

        Kun Qian,\textsuperscript{\rm 1}
        Wei Wei, \textsuperscript{\rm 2}
        Zhou Yu \textsuperscript{\rm 1} \\
}
\affiliations {
    \textsuperscript{\rm 1} University of California, Davis \\
    \textsuperscript{\rm 2} Google Inc.\\
    kunqian@ucdavis.edu, wewei@google.com, joyu@ucdavis.edu
}

\begin{document}

\maketitle

\begin{abstract}
Numerous new dialog domains are being created every day while collecting data for these domains is extremely costly since it involves human interactions.
Therefore, it is essential to develop algorithms that can adapt to different domains efficiently when building data-driven dialog models.
Most recent research on domain adaption focuses on giving the model a better initialization, rather than optimizing the adaptation process.
We propose an efficient domain adaptive task-oriented dialog system model, which incorporates a meta-teacher model to emphasize the different impacts between generated tokens with respect to the context. 
We first train our base dialog model and meta-teacher model adversarially in a meta-learning setting on rich-resource domains. 
The meta-teacher learns to quantify the importance of tokens under different contexts across different domains. 
During adaptation, the meta-teacher guides the dialog model to focus on important tokens in order to achieve better adaptation efficiency.
We evaluate our model on two multi-domain datasets, MultiWOZ and Google Schema-Guided Dialogue, and achieve state-of-the-art performance.
\end{abstract}


\section*{Introduction}
Intelligent personal assistants, such as Amazon's Alexa, have become popular for handling certain simple tasks, such as booking a restaurant, querying bus schedules, and ordering a taxi. 
Often each of these tasks requires a separate task-oriented dialog system to train. Various new domains, such as booking a haircut, a spa, etc., are added every day, so training a dialog system needs thousands of dialogs to leverage the power of deep learning. 
However, the data collection of these tasks is expensive, as humans must be involved.
Therefore, building a task-oriented dialog model that can easily adapt to a new domain with limited data is essential for extending the usability of data-driven dialog systems.


Various methods have been proposed to tackle domain adaptation.
One method is to learn a domain-agnostic embedding space, which represents the semantic meaning of dialog sentences that helps the dialog generation model generalize to new domains~\citep{zhao2018zero}.
Another approach is to take advantage of large pre-trained models, like BERT~\citep{Devlin2019BERTPO}.
Meta-learning methods can be applied for domain adaptation as well.  \citet{qian2019domain} showed the model-agnostic meta-learning  (MAML)~\citep{finn2017model} is promising for extracting domain-specific features during adaptation, which therefore helps the dialog model adapt to the new domain. 
However, all methods above focus on generating a decent initialized model before adaptation and ignore improving the model's adaptive efficiency in the adaptation process. 
To improve the adaptation process and utilize the adaptation data more efficiently, we propose a domain adaptive dialog model based on a meta-learning method and a teacher-student architecture.

During the adaptation step, when the dialog model generates a sequence of tokens as a response, we normally average the loss of all the tokens as the total loss of the entire response.
However, different tokens have different importance with respect to different contexts in different domains.
For example, in the sentence ``what food type do you like?" from the restaurant domain, the tokens ``food type" are more related to this domain compared with other tokens in this sentence.
Therefore, we should pay more attention to this token during adaptation.  If the model fails to generate this token, it should receive more penalty, i.e. larger loss.
Therefore, the focus of our work is utilizing the meta-teacher model to learn each token's weight concerning different contexts for more effective adaptation.

In this paper, we present a \textbf{D}omain \textbf{A}daptive task-oriented dialog model  with \textbf{S}tudent-\textbf{T}eacher architecture (DAST).
We employ the state-of-the-art end-to-end dialog model DAMD~\citep{zhang2019task} as the student model and adopt a transformer-based model as the teacher model. 
The final objective function combines the sequence of token losses generated from the student model and the corresponding weights from the meta-teacher model.
We train these two models adversarially under a model-agnostic meta-learning setting, MAML~\citep{finn2017model}. 
When we adapt the student model to a new domain, we update its parameters with the weighted loss, and fix the parameters of the meta-teacher model during this step.
We evaluate the DAST on two multi-domain task-oriented dialog datasets,  MultiWOZ~\citep{budzianowski2018multiwoz} and  Schema-Guided Dialog~\citep{rastogi2019towards}. 
Experimental results show that our model is effective in extracting domain-specific features and achieves a better domain adaptation performance. 
We will release the code base upon acceptance.


\section*{Related Work}
    
\subsection*{End-to-End Task-Oriented Dialog Systems} 
Traditional task-oriented dialog systems consist of four modules: natural language understanding~\citep{Dauphin2014ZeroShotLA}, dialog state tracker~\citep{Henderson2014TheTD}, dialog policy learning~\citep{Young2010TheHI}, and natural language generation~\citep{Wen2015StochasticLG}. 
Along with the rise of neural networks, more works have explored combing these modules into a single end-to-end model~\citep{wen2016network}.
\citet{lei2018sequicity} proposes to construct the end-to-end dialog model based on a two-stage CopyNet~\citep{Gu2016IncorporatingCM}, which generates belief spans and the delexicalized response~\citep{Wen2015StochasticLG} at the same time.
Moreover, \citet{zhang2019task} incorporate a dialog act predictor and conducts multi-task training~\citep{Li2019EndtoEndTN}. 
The decoder takes in the dialog context and the predicted dialog action together to generate responses.

\subsection*{Domain Adaptation for Dialog System}
Models that can utilize rich-resource domain data to adapt to new low-resource domains effectively have received much attention. 
\citet{Mo2018PersonalizingAD} and \citet{Genevay2016TransferLF} adopt transfer learning method to build a user adaptive dialog model.
\citet{Shi2018SentimentAE} introduces an end-to-end dialog model based on Hybrid Code Netword~\citep{Williams2017HybridCN} for sentiment adaptation.
As for task-oriented dialog systems, \citet{zhao2018zero} and \citet{shalyminov2019few} adopt the typical transfer learning method~\citep{caruana1997multitask,bengio2012deep} and learn latent variables in a domain-agnostic embedding space to solve the problem.
Pre-trained models like BERT~\citep{Devlin2019BERTPO} and GPT-2~\citep{radford2019language} are also introduced to provide decent initialized model parameters for adaption~\citep{budzianowski2019hello,Shalyminov2020HybridGT}. 
The pre-trained model improves the quality and diversity of generated sentences. 
Furthermore, \citet{Peng2020SOLOISTFT} combines a pre-trained model and human correction mechanism, achieving impressive results.

\subsection*{Meta-Learning} 
Meta-learning has been shown to be promising in solving various tasks~\citep{gu2018meta, Rusu2019MetaLearningWL}.
\citet{qian2019domain} applies the MAML~\citep{finn2017model} algorithm to an end-to-end dialog system and demonstrates its decent performance for dialog domain adaptation.
\citet{Song2019LearningTC} modifies MAML for the dialog generation model and learns to customize the model structure for new domains. 
Rather than these previous models which focus on generating an initialized model for adaptation, we pay more attention to improving the process of adaptation. 
We adopt the MAML algorithm to train a meta-teacher model to generate weights to be multiplied with token losses. The meta-teacher learns to distinguish the tokens that the student model needs to focus on from source domains. During adaptation, the meta-teacher instructs the student on which tokens require more attention by assigning weights to tokens and re-weighting each token loss.

                \begin{figure*}[ht!]
                \centering
                \includegraphics[width=0.95\linewidth]{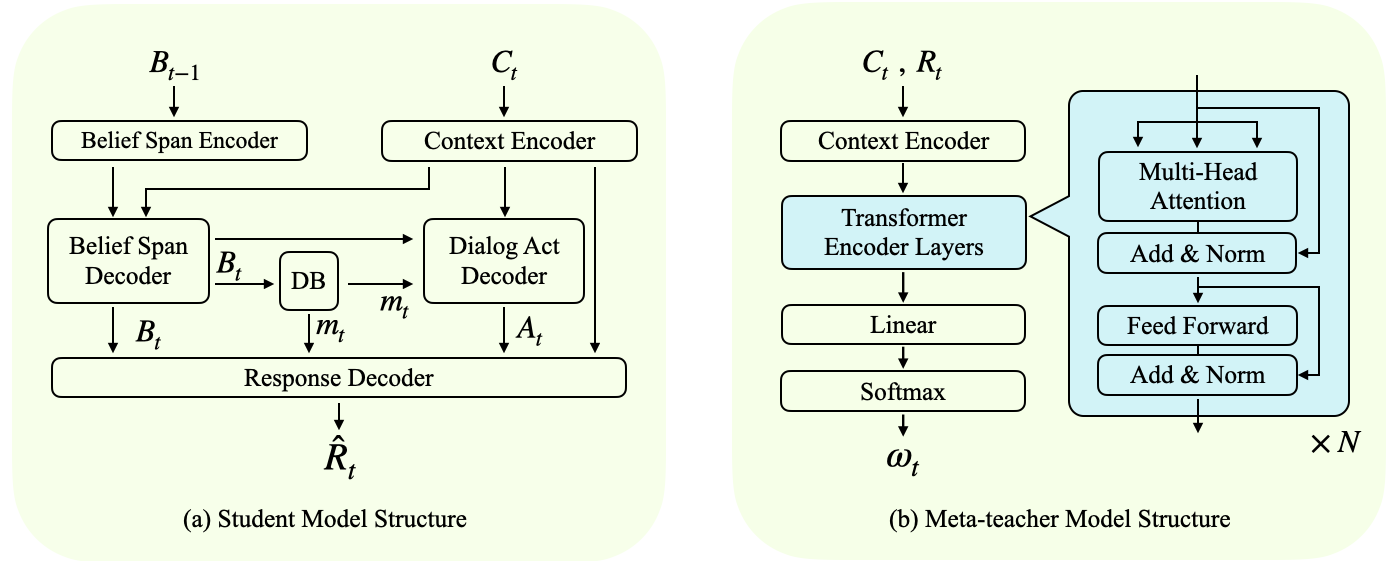}
                \caption{DAST's model architecture. (a) shows the student model's architecture. The input $B_{t-1}$ and $C_t$ represents the previous belief span and dialog context. The student model outputs the current belief span $B_t$, dialog act $A_t$ and system response $\hat{R}_t$. $m_t$ means the number of available choices found in the database (DB) constrained by $B_t$.  (b) illustrates the meta-teacher model's architecture. The meta-teacher takes the context $C_t$ and ground truth system response ${R}_t$ as the input and produces the weights $\omega_t$ as the output.}
                \label{fig_structure}
                \end{figure*}

\subsection*{Teacher-Student Architecture} 
%
The teacher model has been introduced to improve the training process of the student model in various aspects.
Much previous work on teacher model focuses on selecting training data~\citep{Zhu2015MachineTA, Liu2017IterativeMT}. 
\citet{Kim2016SequenceLevelKD} and \citet{Peng2019TeacherStudentFE} distill external knowledge from teacher model to guide the training of student model, based on the idea of knowledge distillation~\citep{Hinton2015DistillingTK},
Moreover, \citet{Fan2018LearningTT} proposes a reinforcement learning-based teacher model, learning to teach (L2T), that can not only teach the model to select data but also select loss function.
Inspired by L2T's work on selecting loss function, \citet{Wu2018LearningTT} constructs a teacher model that simulates different optimal loss functions according to different training stages. 
More specifically, it assigns weights to each token class for machine translation task.
However, with different contexts, even the same token has varying impacts on model training, especially in different domains. 
Therefore we propose a meta-teacher model to generate different weights for tokens with respect to different specific domains and contexts, which instructs the student model where to pay attention to and enables the student model to adapt to a new domain more efficiently.
Since our meta-teacher model focuses on the adaptation process, it is also compatible with other domain adaptation methods.

\section*{Problem Formulation}
We formulate our problem as a  domain adaptation problem. We have $K$ source domains $S_k\ (k=1,2,\dots,K)$ with the corresponding data:
    $$D_{S_k}=\{(c^k_n,r^k_n, S_k), n=1,2,\ \dots\ ,N\}$$
and a target domain $T$ with a limited amount of data for adaptation:
    $$D_T=\{(c^T_n,r^T_n, T), n=1,2,\ \dots\ ,N'\}$$
where $N'\ll N$. Specifically, we set $N$ to be around 50 times larger than $N'$.
In the above equations, $c$ represents the dialog context, which is the dialog model's input. And $r$ refers to the system response, which is the dialog model's output.

There are two stages in the domain adaptation process: training and adaptation.
In the training stage, we learn a general dialog model $\mathcal{M}_{source}$ using all the source domain data. This model is not an optimal model in each individual source domain. However, it can be easily adapt to a well-performed model $\mathcal{M}_{S_k}$ in specific domain through fine-tuning on this domain's data, $D_{S_k}$.
$$\mathcal{M}_{S_k}:C^{S_k}\times S_k\rightarrow R^{S_k}$$
where $C^{S_k}, R^{S_k}$ are the sets of contexts and the system responses from domain $S_k$.
Then, we adapt the model, $\mathcal{M}_{source}$ to the target domain by fine-tuning it with the target domain data, $D_T$. Finally, we obtain a new dialog model $\mathcal{M}_{T}$. 
    $$\mathcal{M}_{T}:C^T\times T\rightarrow R^T$$

\section*{Proposed Method}
In this section, we first describe the detailed structure of DAST in Figure~\ref{fig_structure}, including a student model and a meta-teacher model. We also introduce the forward-propagation process in the first two subsections.
Then we describe student and meta-teacher's architecture and DAST model's training process.


\subsection*{Student Model}
We show the student model in Figure~\ref{fig_structure}(a). It is constructed based on DAMD dialog model~\citep{zhang2019task}, which consists of two encoders and three decoders.
    
Because natural sentences and belief span~\citep{lei2018sequicity} (records all the task-related information provided by user, e.g ``[restaurant] food Chinese price moderate'') have different structures, we build  separate encoders to encode them. $Encoder_B$ is used to encode belief span $B_t$ and $Encoder_C$ is used to encode dialog context $C_t$, which includes the previous responses $R_{t-1}$ and  the current user utterance $U_t$:
            $$h_B=Encoder_B(B_{t-1})$$
            $$h_C=Encoder_C(C_t)$$
We input the hidden state of the previous belief span $h_B$ and context $h_C$ into the belief span decoder to update the belief span:
            $${B}_t=Decoder_B(h_B,h_C)$$
Searching the database with the current belief span, the model generates a one-hot vector $m_t$ to suggest the number of matched entities. 
The dialog act decoder then takes the result obtained from database search, $m_t$, the belief span $B_t$, and the context $C_t$ as input to predict the next system dialog act $A_t$:
            $${A}_t=Decoder_A(B_t,m_t,h_C)$$
Since the model is trained with the data from multiple domains and each domain involves a different number of dialog acts, we use a decoder rather than a classifier to obtain dialog act results.
    
In the end, the response decoder generates system response $\hat{R}_t$ based on all the internal variables ($B_t,m_t,A_t$) and the context $C_t$:
            $$\hat{R}_t=Decoder_R(B_t,m_t,A_t,h_C)$$
We adopt cross-entropy as the basic loss function.
For simplicity, we use GRU~\citep{cho2014learning} with attention layer~\citep{bahdanau2014neural} and copy mechanism~\citep{Gu2016IncorporatingCM} as encoders and decoders.


\subsection*{Meta-Teacher Model}
Figure~\ref{fig_structure}(b) describes the structure of our meta-teacher model. We named the teacher model, meta-teacher as it has no knowledge of the target domain, but has access to samples from all the source domains.
The meta-teacher learns to recognize domain-specific features in a new domain by comparing target domain data with source domain data. Then the meta-teacher model guides the student to focus on unique features in a target domain.

As illustrated in Figure~\ref{fig_structure}(b), we first encode the dialog context $C_t$ and the ground truth system response $R_t$ with the same context encoder $Encoder_C$ as used in the student model.
$$h_1=Encoder_C([C_t,R_t])$$
Then we input the hidden state $h_1$ into multiple transformer~\citep{vaswani2017attention} encoder layers. In our experiment, we set the layer number as two.
$$h_2=TransformerEncoderLayers(h_1)$$
We do not use a complete transformer here because we find that the previous decoded weight does not affect decoding the current weight, which is different from decoding tokens. The encoder layer is enough to generate well-performed weights.

At the end, we map the hidden state $h_2$ to weights with a linear layer and normalize the sequence of weights with a softmax layer:
$$\omega_t= softmax(W\cdot h_2)$$

                            \begin{figure}
                            \centering
                            \includegraphics[width=0.95\columnwidth]{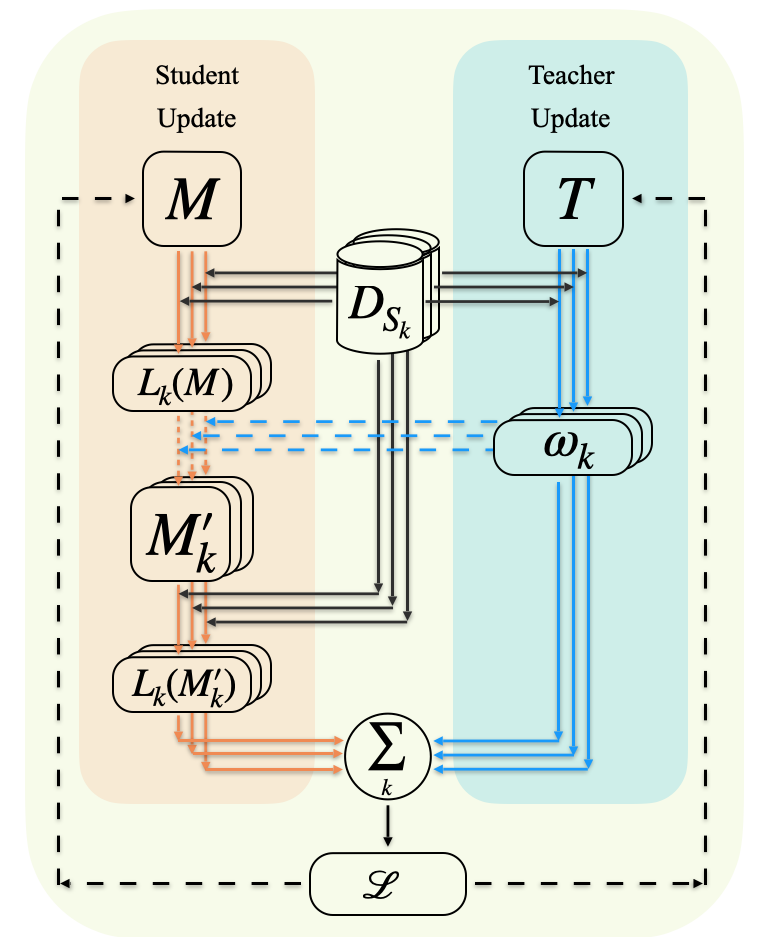}
                            \caption{DAST's training process. Teacher and student model update simultaneously. The solid lines represent the forward propagation steps and the dashed lines represent back-propagation steps}
                            \label{fig_train}
                            \end{figure}

\subsection*{Training Process}

\begin{algorithm}[t] 
\caption*{\textbf{Algorithm} DAST} 
\label{alg1} 
\begin{algorithmic} 
    \renewcommand{\algorithmicrequire}{\textbf{Input:}}
    \REQUIRE Source domain data $\{D^{S_k}\}$; $\alpha$; $\beta$; $\gamma$
    \renewcommand{\algorithmicensure}{\textbf{Output:}} 
    \ENSURE Optimal student and meta-teacher model
        \STATE Randomly initialize student model $\mathcal{M}$ and meta-teacher model $\mathcal{T}$
        \WHILE{$not \ \ converged$}
        \FOR{$S_k\in Source\ \ Domains$}
        \STATE Sample data $(c^k, {r}^k)$ from $D^{S_k}$
        \STATE $\omega_k = \mathcal{T}(c^k, {r}^k)$
        \STATE $Loss_k(\mathcal{M})={\mathit{L}(\mathcal{M'}(c^k),{r}^k)}^T\cdot \omega_k$
        \STATE $\mathcal{M}'=\mathcal{M}-\alpha \nabla_{\mathcal{M}}Loss_k(\mathcal{M})$
        \STATE $Loss_k(\mathcal{M'})={\mathit{L}(\mathcal{M'}(c^k),{r}^k)}^T\cdot \omega_k$
    \ENDFOR
    
    $\mathcal{M} \leftarrow \mathcal{M} - \beta \nabla_{\mathcal{M}} \sum_{k} Loss_k(\mathcal{M}')$\\
    $\mathcal{T} \leftarrow \mathcal{T} + \gamma \nabla_{\mathcal{T}} \sum_{k} Loss_k(\mathcal{M}')$
    \ENDWHILE
    \STATE 
    \renewcommand{\algorithmicrequire}{\textbf{Student Model}}
    \REQUIRE $\mathcal{M}(c=\{B_{t-1},C_t\}):$
  
        \STATE $h_B, h_C = Encoder_B(B_{t-1}), Encoder_C(C_t)$
        \STATE $B_t = Decoder_B(h_B, h_C)$
        \STATE $m_t = SeachDatabase(B_t)$
        \STATE $A_t = Decoder_A(h_C, B_t, m_t)$
        \STATE $R_t = Decoder_R(h_C,B_t,m_t, A_t)$

        \STATE \textbf{return} $R_t$  
\end{algorithmic}
\end{algorithm}

Figure~\ref{fig_train} displays the simultaneous training process of the student and the meta-teacher model under the MAML~\citep{finn2017model} setting. MAML adopts a two-step updating strategy so that the learned model parameter is easy to adapt to a new domain. We apply MAML on both student and meta-teacher models.
    
We first initialize the student and the meta-teacher model's parameters $M$ and $T$ separately. 
For each source domain $S_k$, a data pair $(c^k,{r}^k)$ is sampled from dataset $D_{S_k}$. 
Taking in the data pair, the meta-teacher model generates a  vector of weights $\omega_k$, corresponding to tokens in the true response ${r}^k$.
Meanwhile, the student model computes the cross entropy loss ${\mathit{L}(\mathcal{M}(c^k)_i,{r}_i^k)}$ of each generated token, where $i\in\{1,2,\dots,|{r}^k|\}$. 
After weighting each token's loss, we obtain the final loss:
$$Loss(\mathcal{M}( c^k),{r}^k,\omega_k)={\mathit{L}(\mathcal{M}(c^k),{r}^k)}^T\cdot \omega_k$$
According to MAML, we then temporarily update the student model with gradient descent:
$$\mathcal{M}'=\mathcal{M}-\alpha \nabla_{\mathcal{M}}Loss(\mathcal{M}(c^k), {r}^k, \omega_k)$$
and compute the loss with the updated model similarly:

$$Loss(\mathcal{M'}(c^k),{r}^k,\omega_k)={\mathit{L}(\mathcal{M'}(c^k),{r}^k)}^T\cdot \omega_k$$

Here, for simplicity, we use the same data pair for both updates. However, we update weights when a new data pair is sampled to compute the second loss.
We combine all loss values for all the sampled source domains to obtain the final loss function:
\begin{equation}\label{eq3}
\mathscr{L}(\mathcal{M},\mathcal{T}, D_S)=\sum_k \mathit{Loss}(\mathcal{M}'( c^k), {r}^k, \omega_k)
\end{equation}
Finally, we optimize the student model by minimizing the final loss:
$$\mathcal{M} \leftarrow \mathcal{M} - \beta \nabla_{\mathcal{M}} \mathscr{L}(\mathcal{M},\mathcal{T}, D_S)$$
The teacher model's responsibility is to guide the student model to extract domain-specific features and quickly adapt to a target domain. 
This means the meta-teacher should give large weights to domain-related tokens. Since such tokens do not show up frequently in source domains, the student model does not perform well when predicting these tokens, which produces a large token loss.
So meta-teacher's weight should maximize the final loss in Eq.~(\ref{eq3}).
Consequently, we train the meta-teacher model and student model adversarially by optimizing:
$$\min_{\mathcal{M}}\max_{\mathcal{T}}\mathscr{L}(\mathcal{M},\mathcal{T}, D_S)$$
Optionally, a regularization term can be added to the loss function $\mathscr{L}(\mathcal{M},\mathcal{T}, D_S)$ to avoid situations where all weights are the same. In our experiments, we adopt the L2 regularization term.

Besides, we update the meta-teacher model during both training and validation stages. Since the student model cannot observe validation data, this does not affect the student model to learn when to stop training. In addition, training with more data enables the meta-teacher to learn the domains comprehensively and distinguish domain-specific features better.

During the adaptation stage, we still compute the weights with the meta-teacher model and multiply the weights with token losses for more efficient adaptation.


\section*{Experiments}
We first introduce the datasets and the metrics used to evaluate DAST. Then, we describe the baseline models used to compare against our model. Finally, we describe the implementation details, including hyper-parameters, of our model.

\subsection*{Datasets}
\label{multiwoz}
\noindent \textbf{MultiWOZ}~\citep{budzianowski2018multiwoz} is a human-human task-oriented dialog dataset, covering seven domains. Since certain dialog covers multiple domains, we extract only single-domain dialogs and each domain averagely contains 487 dialogs. Once we select one domain as the target domain, we adopt all the data from the other six domains as training data. For the adaptation, we randomly choose nine dialogs ($2\%$ of source domain) in the target domain as adaptation data and leave the rest for testing. For each experiment, to reduce randomness in the few-shot learning setting, we repeat the adaptation process for 10 times and report the average result. 

\noindent\textbf{Schema-Guided Dataset}~\citep{rastogi2019towards} is another newly-released human-human dataset of task-oriented dialogs. It was collected in the Wizard-Of-Oz style~\citep{Kelley1984AnID}. The dataset consists of 20 topics, each of which contains multiple tasks (e.g. both task ``looking for a dentist" and task ``looking for a salon" belongs to ``Services" topic) and we consider each task as a domain. Then, we filter out the domains that do not require the dialog system to provide an entity to complete the tasks and the domains that contain less than 100 dialogs. There are 21 domains from 12 topics left. We randomly select seven domains from different topics as source domains for training and randomly select one domain from each of the other topics as target domains. Similar to the setting on MultiWOZ dataset, we randomly choose nine dialogs for adaptation in each domain and test our model on the other dialogs. We report the average result of each metric over 10 runs.


\subsection*{Evaluation Metrics}

Following \citet{budzianowski2018multiwoz}, we adopt Inform rate, Success rate, and BLEU~\citep{Papineni2002BleuAM} score as our main metrics. 

\noindent\textbf{Inform rate} represents the accuracy of successfully providing the correct entity (e.g. the name of a restaurant that satisfies all user's constraints in the restaurant domain). 

\noindent\textbf{Success rate} measures how much the system answers all the requested information. 

\noindent\textbf{BLEU score} is adopted to evaluate the quality of the generated response, compared with the oracle human response.

In addition, we also report the Slot F1 score and the Act F1 score. The Slot F1 checks if the belief span decoder correctly extracts and generates belief states, while the Act F1 measures the matching of predicted dialog act, generated by the dialog act decoder.

\begin{table*}[ht]
\centering
\setlength{\extrarowheight}{0.06cm}
\small
\begin{tabular}[width=\textwidth]{lC{0.75cm}C{0.75cm}C{0.75cm}C{0.75cm}C{0.75cm}C{0.75cm}C{0.75cm}C{0.75cm}C{0.75cm}C{0.75cm}C{0.75cm}c}
\toprule
\hline
& \multicolumn{3}{c}{Attraction} & \multicolumn{3}{c}{Restaurant} & \multicolumn{3}{c}{Train} & \multicolumn{3}{c}{Hotel} \\
\cmidrule(lr){2-4} \cmidrule(lr){5-7} \cmidrule(lr){8-10} \cmidrule(lr){11-13}
Model   & Inform    & Success   & BLEU   & Inform    & Success   & BLEU   & Inform  & Success  & BLEU & Inform  & Success  & BLEU \\ 
\midrule
MAML  & 45.5                 & 22.5                 & 9.5                  & 46.0                 & 10.8                 & 7.0                  & 76.3                 & 49.0                 & \textbf{5.7}                  & 48.6                 & 17.7                 & 6.6                    \\
DAST   & \textbf{54.7}                 & \textbf{32.5}                 & \textbf{10.2 }                & \textbf{51.2}                 & \textbf{17.5 }                & \textbf{8.0 }                 & \textbf{76.9  }               & \textbf{50.0 }                & 5.5                  & \textbf{49.1 }                & \textbf{25.1}                 & \textbf{7.6 }                \\
\hline
\hline
& \multicolumn{3}{c}{Taxi} & \multicolumn{3}{c}{Police} & \multicolumn{3}{c}{Hospital} & \multicolumn{3}{c}{Average} \\
\cmidrule(lr){2-4} \cmidrule(lr){5-7} \cmidrule(lr){8-10} \cmidrule(lr){11-13}
Model   & Inform    & Success   & BLEU   & Inform    & Success   & BLEU   & Inform  & Success  & BLEU & Inform  & Success  & BLEU \\ 
\midrule
MAML    & -                & 59.8                 & 7.7                  & -              & 41.2                 & 7.0                  & -                & 47.6                 & 9.8                  & 54.1                 & 35.5                 & 7.6 \\
DAST    & -                & \textbf{61.8}                 & 7.7                  & -            & \textbf{47.2 }                & \textbf{8.2}                & -                     & \textbf{48.2}                 & \textbf{10.4}                 & \textbf{58.0 }                & \textbf{40.3 }                & \textbf{8.2  }                \\
\bottomrule
\end{tabular}
\caption{The performance in the metrics of Inform rate, Success rate, and BLEU score on all seven domains from MultiWOZ, as well as the average values over domains. We do not report the Inform rate in domain ``Taxi'', ``Police'' and ``Hospital'' because each of these three domains contains a default task entity. 
DAST outperforms the MAML baseline in terms of Inform rate and Success rate in every domain and achieves better average BLEU score.}
\label{multiwoz-main}
\end{table*}

\subsection*{Baselines}
To evaluate the effectiveness of our model, we compare DAST with the following two baselines:


\noindent \textbf{MAML}~\citep{finn2017model} is simple but powerful for domain adaptation. We apply MAML to train the student model and conduct adaptation with gradient descent. This baseline is similar to DAML proposed by \citet{qian2019domain}, except that the dialog model in the baseline adds the belief span encoder and the dialog act decoder. 

\noindent \textbf{SOLOIST}~\citep{Peng2020SOLOISTFT} is a newly-released model that is based on pre-trained GPT-2 model. This model achieves state-of-the-art performance for domain adaptation on the MultiWOZ dataset. However, its code has not been published, we only compare its results with the same settings our model used.

\subsection*{Implementation Details}
We adopt GloVe~\citep{pennington2014glove} as the initialized value for word embeddings, with an embedding size of 50. For the student model, each GRU from encoders and decoders contains one layer and the hidden size is set as 100. Furthermore, the GRU models of two encoders are bi-directional. As for the teacher model, it contains 2 self-attention layers with 5 heads for each. We use Adam~\citep{kingma2014adam} for optimization and set an initialized learning rate as 0.005 for both student and teacher model, as well as the meta optimizer. The learning rate decays by half if no improvement is observed on validation data for 3 successive epochs and the training process would stop early when no improvement is observed on validation data for 5 successive epochs. We adopt the batch normalization~\citep{ioffe2015batch} and use a batch size of 32.


\section*{Results and Analysis}

%
Table~\ref{multiwoz-main} lists the model's results on the metrics of Inform rate, Success rate, and BLEU score in all seven domains from the MultiWOZ dataset. We do not report the Inform rate of ``Taxi'', ``Police'' and ``Hospital'' domain because these three domains have default informable entities, which means the Inform rate is always $100\%$.
%
%

\begin{table}[ht!]
\centering
\setlength{\extrarowheight}{0.06cm}
\small
\begin{tabular}[width=\columnwidth]{l|cccc}
\toprule
\hline
& \multicolumn{2}{c}{Slot F1} & \multicolumn{2}{c}{Act F1}\\
\cmidrule(lr){2-3} \cmidrule(lr){4-5} 
Domains  &MAML& DAST &MAML& DAST  \\ 
\hline
Attraction & 31.5                 & \textbf{38.4}                 & 40.0                 & \textbf{41.4}  \\
\hline
Restaurant & 36.1                 & \textbf{42.5 }                & 32.7                 & \textbf{36.9}  \\
\hline
Train     & 37.5                 & \textbf{41.8}                 & 29.0                 & \textbf{31.0}\\
\hline
Hotel& 22.4                 & \textbf{26.0}                 & 27.2                 & \textbf{29.1} \\
\hline
Taxi & -                & -               & 44.0                 & \textbf{45.2}                  \\
\hline
Police & -                 &-                & 51.5                 & \textbf{53.9 }                 \\
\hline
Hospital&-         & -      & \textbf{52.1}   & 51.1  \\
\hline
\textbf{Average}  & 31.9    & \textbf{37.2}  & 39.5   & \textbf{41.2}\\
\hline
\bottomrule
\end{tabular}
\caption{The evaluation results on the metrics of Slot F1 and Act F1 in all domains. The DAST outperforms the MAML baseline in terms of Slot F1 in every domain and predicts better dialog acts on average. }
\label{multiwoz_slot}
\end{table}

The results in the Table~\ref{multiwoz-main} show that, for each domain, our model outperforms the baselines in terms of both Inform rate and Success rate. This suggests that the weights generated by the meta-teacher model are beneficial for the student model to optimize adaptation process and achieve better performance in dialog task completion.
For the other metric, the BLEU score, our model does not consistently outperform the baselines. This is because the original unweighted loss treats every token in the same way in order to instruct the student model to learn the probabilistic language model. Therefore, the weights from meta-teacher slightly disturb this learning process and consequently reduce the BLEU score. However, our model still achieves better BLEU score than MAML baseline on average, indicating that although slightly affecting learning the language model, the meta-teacher helps the student model to learn new features of the new domain in general.
%

\begin{table}[h]
\centering
\setlength{\extrarowheight}{0.06cm}
\small
\begin{tabular}[width=\columnwidth]{l|C{0.9cm}C{0.85cm}C{0.8cm}C{0.8cm}c}
    \toprule
    \hline
    \textbf{Model}  & Slot F1       &Act F1    & Inform      &Success     & BLEU\\
    \hline
    MAML    &  22.5&50.3&69.2&47.9   &12.0   \\
    DAST    & \textbf{23.2}&\textbf{59.5}&\textbf{89.6}&\textbf{61.6}&\textbf{12.1} \\
    \hline
    \bottomrule

\end{tabular}
\caption{The average performance of all target domains in Schema-Guided Dataset. The DAST outperforms the MAML baseline in terms of all reported metrics}
\label{schema}
\end{table}

The performance on the other two metrics is shown in Table~\ref{multiwoz_slot}.
Since each of ``Taxi'', ``Police'' and ``Hospital'' domain has a default task entity, which means the explicit state tracking is not required to accomplish the task, we do not report the results on Slot F1 in these three domains. On average, our model outperforms the MAML baseline in both Slot F1 and Act F1, suggesting that the meta-teacher also encourages the dialog model to generate the correct dialog states, which is necessary for database search. And only after searching the database with correct constraints, the dialog system can provide user with a correct task entity.


\begin{table*}[ht]
\centering
\setlength{\extrarowheight}{0.08cm}
\small
\begin{tabular}[width=\textwidth]{lC{0.75cm}C{0.75cm}C{0.75cm}C{0.75cm}C{0.75cm}C{0.75cm}C{0.75cm}C{0.75cm}C{0.75cm}C{0.75cm}C{0.75cm}c}
\toprule
\hline
& \multicolumn{3}{c}{$80$} & \multicolumn{3}{c}{$400$} & \multicolumn{3}{c}{$800$} & \multicolumn{3}{c}{$1600$} \\
\cmidrule(lr){2-4} \cmidrule(lr){5-7} \cmidrule(lr){8-10} \cmidrule(lr){11-13}
Model   & Inform    & Success   & BLEU   & Inform    & Success   & BLEU   & Inform  & Success  & BLEU & Inform  & Success  & BLEU \\ 
\midrule
SOLOIST                      & 58.4                 & 35.3                 & \textbf{10.58}                & 69.3                 & 52.3                 & \textbf{11.8}                 & 69.9                 & 51.9                 & \textbf{14.6}                 & 74                   & 60.1                 & \textbf{15.24}                \\
MAML                         & 62.09                & 38.36                & 9.96                     & 72.31                & 52.91                & 10.87                      & 74.78                & 57.71                & 11.29                      & 76.73                & 60.61                & 11.99                      \\
DAST                          & \textbf{62.70}                & \textbf{38.68}                & 9.49                & \textbf{74.52}                & \textbf{54.45}                & 11.08                     & \textbf{75.92}                & \textbf{57.72}                & 11.52                      & \textbf{77.95}                & \textbf{60.87}                & 11.97                     \\
\bottomrule
\end{tabular}
\caption{The average performance over four domains, ``Attraction'', ``Restaurant'', ``Train'' and``Hotel'', with 80/400/800/1600 dialogs for adaptation. DAST consistently achieves the best performance in the metrics of Inform rate and Success rate, with increasing the amount of adaptation data
}
\label{multiwoz_increase_adapt_size}
\end{table*}

We also explore the relationship between model performance and the amount of adaptation data. To keep the same setting as SOLOIST~\citep{Peng2020SOLOISTFT}, we choose only four domains, ``Attraction'', ``Restaurant'', ``Train'' and ``Hotel'' for evaluation. We randomly sample either 80, 400, 800, or 1600 dialogs in total over four domains. We show all model's results in Table~\ref{multiwoz_increase_adapt_size}. We follow SOLOIST and only report Inform rate, Success rate, and BLEU score.
We can find that our method consistently outperforms the baselines, both SOLOIST and MAML, in terms of Inform rate and Success rate. Since these two metrics are closely related to task completion, we believe meta-teacher guides the student model to adapt to new domains more efficiently.
On the other hand, the SOLOIST model performs the best in the BLEU score. One main reason is that SOLOIST is fine-tuned based on a pre-trained language model, GPT-2. The pre-trained model makes SOLOIST generates more fluent sentences. However, our teacher-student architecture is generalizable to different student model structures, and can be built based on pre-trained student model as well.
In addition, we find that the gap between SOLOIST and our methods reduces with an increasing amount of adaptation data. This is because, with enough data, the student model can learn the new domain well without the meta-teacher's guidance. Therefore, the influence of the meta-teacher declines as the number of adaptation dialogs increases.

Table~\ref{schema} describes the average performance in all the target domains from the Schema-Guided dataset. Our method achieves better performance compared to the MAML baseline in all metrics, suggesting that our method can generalize to different multi-domain dialog datasets.


\section*{Case Study and Visualization}
\begin{figure}[t]
\centering
\includegraphics[width=\columnwidth]{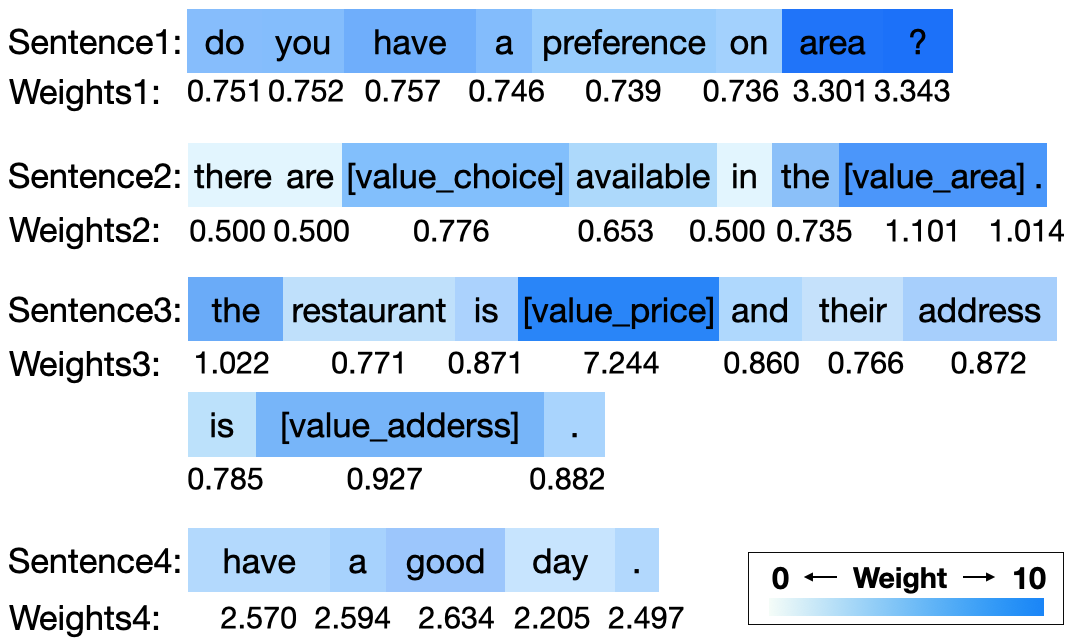}
\caption{Visualizing weights corresponding to different tokens with different color intensities. The darker the color is, the larger the corresponding token's weight is.}
\label{fig_weights}
\end{figure}

Figure~\ref{fig_weights} lists four example sentences in the restaurant domain from  the MultiWOZ dataset, along with their weights assigned by the meta-teacher model. 
To visualize the weight of each token, we color each token according to its corresponding weight. 
The larger the weight is, the darker the color is. 
Since we multiply the weights with token losses to update the student model, the absolute value of the weight can be considered as part of the learning rate. 
Therefore, we mainly focus on the relative values of weights within the same sentence. 
And the color intensity only suggests the relative value of weights in the same sentence.

The first two sentences show that our meta-teacher model focuses more on the domain-related tokens like ``area'' and delexicalized slots such as ``[value$\_$area]''. 
One possible reason is that general tokens (like ``there are'' in the second sentence) have already been learned by the student model in source domains while tokens like ``[value$\_$area]'' appear less frequently in the source domains. 
Since large weight amplifies the feedback of the token loss during back-propagation, larger weights for the domain-related tokens encourage the student model to focus on those tokens and quickly learn features of the new domain, leading to more efficient adaptation.
In the third case, 
we find that the delexicalized slots like ``[value$\_$range]" and ``[value$\_$address]" attracts more attention than domain-related token ``address''.
This is because the domain-related tokens are still possible to be found in other domains. 
For example, ``address'' exists in five domains in the MultiWOZ dataset. 
The last sentence does not contain any domain-specific tokens. Hence, the token weights are close to each other. In this case, the weights do not make much impacts on updating model parameters.

\section*{Conclusion and Future work}

We propose a domain adaptation method for low-resource task-oriented dialog systems, which incorporates a student-teacher architecture under the meta-learning setting.
%
We present a transformer-based meta-teacher model, which learns to distinguish important tokens under different contexts across source domains during training. As for adaptation, the meta-teacher instructs the student dialog model to pay more attention to influential tokens by assigning weights to token losses, which improves the student model's adaptation performance.
We evaluate our method on two popular human-human multi-domain datasets. 
The results demonstrate that our method reaches state-of-the-art performance in most task-related metrics, compared with MAML and SOLOIST. 
Since the meta-teacher is built to assign weights to a sequence of generated tokens, our method can be applied to other NLP tasks, such as machine translation and summarization.
Furthermore, our meta-teacher model is compatible with other domain adaptation methods, such as MAML and pre-trained models.

In the future, we aim to extend our method in several directions. First, we plan to include the Success rate and Inform rate into the loss function of the meta-teacher model in a reinforcement learning setting. We believe directly optimize task success metric may lead to better performance. Another direction is to combine the meta-teacher model and pre-trained models to explore the compatibility, as well as replacing GRU-based student model with pre-trained models for better sentence quality.

\bibliography{aaai21.bib}

\begin{thebibliography}{44}
\providecommand{\natexlab}[1]{#1}
\providecommand{\url}[1]{\texttt{#1}}
\providecommand{\urlprefix}{URL }
\expandafter\ifx\csname urlstyle\endcsname\relax
  \providecommand{\doi}[1]{doi:\discretionary{}{}{}#1}\else
  \providecommand{\doi}{doi:\discretionary{}{}{}\begingroup
  \urlstyle{rm}\Url}\fi

\bibitem[{Bahdanau, Cho, and Bengio(2014)}]{bahdanau2014neural}
Bahdanau, D.; Cho, K.; and Bengio, Y. 2014.
\newblock Neural machine translation by jointly learning to align and
  translate.
\newblock \emph{arXiv preprint arXiv:1409.0473} .

\bibitem[{Bengio(2012)}]{bengio2012deep}
Bengio, Y. 2012.
\newblock Deep learning of representations for unsupervised and transfer
  learning.
\newblock In \emph{Proceedings of ICML Workshop on Unsupervised and Transfer
  Learning}, 17--36.

\bibitem[{Budzianowski and Vuli{\'c}(2019)}]{budzianowski2019hello}
Budzianowski, P.; and Vuli{\'c}, I. 2019.
\newblock Hello, It's GPT-2--How Can I Help You? Towards the Use of Pretrained
  Language Models for Task-Oriented Dialogue Systems.
\newblock \emph{arXiv preprint arXiv:1907.05774} .

\bibitem[{Budzianowski et~al.(2018)Budzianowski, Wen, Tseng, Casanueva, Ultes,
  Ramadan, and Ga{\v{s}}i{\'c}}]{budzianowski2018multiwoz}
Budzianowski, P.; Wen, T.-H.; Tseng, B.-H.; Casanueva, I.; Ultes, S.; Ramadan,
  O.; and Ga{\v{s}}i{\'c}, M. 2018.
\newblock Multiwoz-a large-scale multi-domain wizard-of-oz dataset for
  task-oriented dialogue modelling.
\newblock \emph{arXiv preprint arXiv:1810.00278} .

\bibitem[{Caruana(1997)}]{caruana1997multitask}
Caruana, R. 1997.
\newblock Multitask learning.
\newblock \emph{Machine learning} 28(1): 41--75.

\bibitem[{Cho et~al.(2014)Cho, Van~Merri{\"e}nboer, Gulcehre, Bahdanau,
  Bougares, Schwenk, and Bengio}]{cho2014learning}
Cho, K.; Van~Merri{\"e}nboer, B.; Gulcehre, C.; Bahdanau, D.; Bougares, F.;
  Schwenk, H.; and Bengio, Y. 2014.
\newblock Learning phrase representations using RNN encoder-decoder for
  statistical machine translation.
\newblock \emph{arXiv preprint arXiv:1406.1078} .

\bibitem[{Dauphin et~al.(2014)Dauphin, T{\"u}r, Hakkani-T{\"u}r, and
  Heck}]{Dauphin2014ZeroShotLA}
Dauphin, Y.; T{\"u}r, G.; Hakkani-T{\"u}r, D.~Z.; and Heck, L.~P. 2014.
\newblock Zero-Shot Learning and Clustering for Semantic Utterance
  Classification.
\newblock \emph{CoRR} abs/1401.0509.

\bibitem[{Devlin et~al.(2019)Devlin, Chang, Lee, and
  Toutanova}]{Devlin2019BERTPO}
Devlin, J.; Chang, M.-W.; Lee, K.; and Toutanova, K. 2019.
\newblock BERT: Pre-training of Deep Bidirectional Transformers for Language
  Understanding.
\newblock \emph{ArXiv} abs/1810.04805.

\bibitem[{Fan et~al.(2018)Fan, Tian, Qin, Li, and Liu}]{Fan2018LearningTT}
Fan, Y.; Tian, F.; Qin, T.; Li, X.; and Liu, T.-Y. 2018.
\newblock Learning to Teach.
\newblock \emph{ArXiv} abs/1805.03643.

\bibitem[{Finn, Abbeel, and Levine(2017)}]{finn2017model}
Finn, C.; Abbeel, P.; and Levine, S. 2017.
\newblock Model-agnostic meta-learning for fast adaptation of deep networks.
\newblock In \emph{Proceedings of the 34th International Conference on Machine
  Learning-Volume 70}, 1126--1135. JMLR. org.

\bibitem[{Genevay and Laroche(2016)}]{Genevay2016TransferLF}
Genevay, A.; and Laroche, R. 2016.
\newblock Transfer Learning for User Adaptation in Spoken Dialogue Systems.
\newblock In \emph{AAMAS}.

\bibitem[{Gu et~al.(2016)Gu, Lu, Li, and Li}]{Gu2016IncorporatingCM}
Gu, J.; Lu, Z.; Li, H.; and Li, V. O.~K. 2016.
\newblock Incorporating Copying Mechanism in Sequence-to-Sequence Learning.
\newblock \emph{CoRR} abs/1603.06393.

\bibitem[{Gu et~al.(2018)Gu, Wang, Chen, Cho, and Li}]{gu2018meta}
Gu, J.; Wang, Y.; Chen, Y.; Cho, K.; and Li, V.~O. 2018.
\newblock Meta-learning for low-resource neural machine translation.
\newblock \emph{arXiv preprint arXiv:1808.08437} .

\bibitem[{Henderson, Thomson, and Williams(2014)}]{Henderson2014TheTD}
Henderson, M.; Thomson, B.; and Williams, J.~D. 2014.
\newblock The third Dialog State Tracking Challenge.
\newblock \emph{2014 IEEE Spoken Language Technology Workshop (SLT)} 324--329.

\bibitem[{Hinton, Vinyals, and Dean(2015)}]{Hinton2015DistillingTK}
Hinton, G.~E.; Vinyals, O.; and Dean, J. 2015.
\newblock Distilling the Knowledge in a Neural Network.
\newblock \emph{ArXiv} abs/1503.02531.

\bibitem[{Ioffe and Szegedy(2015)}]{ioffe2015batch}
Ioffe, S.; and Szegedy, C. 2015.
\newblock Batch normalization: Accelerating deep network training by reducing
  internal covariate shift.
\newblock \emph{arXiv preprint arXiv:1502.03167} .

\bibitem[{Kelley(1984)}]{Kelley1984AnID}
Kelley, J.~F. 1984.
\newblock An iterative design methodology for user-friendly natural language
  office information applications.
\newblock \emph{ACM Trans. Inf. Syst.} 2: 26--41.

\bibitem[{Kim and Rush(2016)}]{Kim2016SequenceLevelKD}
Kim, Y.; and Rush, A.~M. 2016.
\newblock Sequence-Level Knowledge Distillation.
\newblock \emph{ArXiv} abs/1606.07947.

\bibitem[{Kingma and Ba(2014)}]{kingma2014adam}
Kingma, D.~P.; and Ba, J. 2014.
\newblock Adam: A method for stochastic optimization.
\newblock \emph{arXiv preprint arXiv:1412.6980} .

\bibitem[{Lei et~al.(2018)Lei, Jin, Kan, Ren, He, and Yin}]{lei2018sequicity}
Lei, W.; Jin, X.; Kan, M.-Y.; Ren, Z.; He, X.; and Yin, D. 2018.
\newblock Sequicity: Simplifying task-oriented dialogue systems with single
  sequence-to-sequence architectures.
\newblock In \emph{Proceedings of the 56th Annual Meeting of the Association
  for Computational Linguistics (Volume 1: Long Papers)}, 1437--1447.

\bibitem[{Li et~al.(2019)Li, Qian, Shi, and Yu}]{Li2019EndtoEndTN}
Li, Y.; Qian, K.; Shi, W.; and Yu, Z. 2019.
\newblock End-to-End Trainable Non-Collaborative Dialog System.
\newblock \emph{ArXiv} abs/1911.10742.

\bibitem[{Liu et~al.(2017)Liu, Dai, Humayun, Tay, Yu, Smith, Rehg, and
  Song}]{Liu2017IterativeMT}
Liu, W.; Dai, B.; Humayun, A.; Tay, C.; Yu, C.; Smith, L.~B.; Rehg, J.~M.; and
  Song, L. 2017.
\newblock Iterative Machine Teaching.
\newblock \emph{ArXiv} abs/1705.10470.

\bibitem[{Mo et~al.(2018)Mo, Zhang, Li, Li, and Yang}]{Mo2018PersonalizingAD}
Mo, K.; Zhang, Y.; Li, S.; Li, J.; and Yang, Q. 2018.
\newblock Personalizing a Dialogue System With Transfer Reinforcement Learning.
\newblock In \emph{AAAI}.

\bibitem[{Papineni et~al.(2002)Papineni, Roukos, Ward, and
  Zhu}]{Papineni2002BleuAM}
Papineni, K.; Roukos, S.; Ward, T.; and Zhu, W.-J. 2002.
\newblock Bleu: a Method for Automatic Evaluation of Machine Translation.

\bibitem[{Peng et~al.(2020)Peng, Li, chao Li, Shayandeh, Liden, and
  Gao}]{Peng2020SOLOISTFT}
Peng, B.; Li, C.; chao Li, J.; Shayandeh, S.; Liden, L.; and Gao, J. 2020.
\newblock SOLOIST: Few-shot Task-Oriented Dialog with A Single Pre-trained
  Auto-regressive Model.
\newblock \emph{ArXiv} abs/2005.05298.

\bibitem[{Peng et~al.(2019)Peng, Huang, Lin, Ji, Chen, and
  Zhang}]{Peng2019TeacherStudentFE}
Peng, S.; Huang, X.; Lin, Z.; Ji, F.; Chen, H.; and Zhang, Y. 2019.
\newblock Teacher-Student Framework Enhanced Multi-domain Dialogue Generation.
\newblock \emph{ArXiv} abs/1908.07137.

\bibitem[{Pennington, Socher, and Manning(2014)}]{pennington2014glove}
Pennington, J.; Socher, R.; and Manning, C.~D. 2014.
\newblock GloVe: Global Vectors for Word Representation.
\newblock In \emph{Empirical Methods in Natural Language Processing (EMNLP)},
  1532--1543.
\newblock \urlprefix\url{http://www.aclweb.org/anthology/D14-1162}.

\bibitem[{Qian and Yu(2019)}]{qian2019domain}
Qian, K.; and Yu, Z. 2019.
\newblock Domain Adaptive Dialog Generation via Meta Learning.
\newblock \emph{arXiv preprint arXiv:1906.03520} .

\bibitem[{Radford et~al.(2019)Radford, Wu, Child, Luan, Amodei, and
  Sutskever}]{radford2019language}
Radford, A.; Wu, J.; Child, R.; Luan, D.; Amodei, D.; and Sutskever, I. 2019.
\newblock Language models are unsupervised multitask learners.
\newblock \emph{OpenAI Blog} 1(8): 9.

\bibitem[{Rastogi et~al.(2019)Rastogi, Zang, Sunkara, Gupta, and
  Khaitan}]{rastogi2019towards}
Rastogi, A.; Zang, X.; Sunkara, S.; Gupta, R.; and Khaitan, P. 2019.
\newblock Towards scalable multi-domain conversational agents: The
  schema-guided dialogue dataset.
\newblock \emph{arXiv preprint arXiv:1909.05855} .

\bibitem[{Rusu et~al.(2019)Rusu, Rao, Sygnowski, Vinyals, Pascanu, Osindero,
  and Hadsell}]{Rusu2019MetaLearningWL}
Rusu, A.~A.; Rao, D.; Sygnowski, J.; Vinyals, O.; Pascanu, R.; Osindero, S.;
  and Hadsell, R. 2019.
\newblock Meta-Learning with Latent Embedding Optimization.
\newblock \emph{ArXiv} abs/1807.05960.

\bibitem[{Shalyminov et~al.(2019)Shalyminov, Lee, Eshghi, and
  Lemon}]{shalyminov2019few}
Shalyminov, I.; Lee, S.; Eshghi, A.; and Lemon, O. 2019.
\newblock Few-Shot Dialogue Generation Without Annotated Data: A Transfer
  Learning Approach.
\newblock \emph{arXiv preprint arXiv:1908.05854} .

\bibitem[{Shalyminov et~al.(2020)Shalyminov, Sordoni, Atkinson, and
  Schulz}]{Shalyminov2020HybridGT}
Shalyminov, I.; Sordoni, A.; Atkinson, A.; and Schulz, H. 2020.
\newblock Hybrid Generative-Retrieval Transformers for Dialogue Domain
  Adaptation.
\newblock \emph{ArXiv} abs/2003.01680.

\bibitem[{Shi and Yu(2018)}]{Shi2018SentimentAE}
Shi, W.; and Yu, Z. 2018.
\newblock Sentiment Adaptive End-to-End Dialog Systems.
\newblock In \emph{ACL}.

\bibitem[{Song et~al.(2019)Song, Liu, Bi, Yan, and Zhang}]{Song2019LearningTC}
Song, Y.; Liu, Z.; Bi, W.; Yan, R.; and Zhang, M. 2019.
\newblock Learning to Customize Model Structures for Few-shot Dialogue
  Generation Tasks.

\bibitem[{Vaswani et~al.(2017)Vaswani, Shazeer, Parmar, Uszkoreit, Jones,
  Gomez, Kaiser, and Polosukhin}]{vaswani2017attention}
Vaswani, A.; Shazeer, N.; Parmar, N.; Uszkoreit, J.; Jones, L.; Gomez, A.~N.;
  Kaiser, {\L}.; and Polosukhin, I. 2017.
\newblock Attention is all you need.
\newblock In \emph{Advances in neural information processing systems},
  5998--6008.

\bibitem[{Wen et~al.(2015)Wen, Gasic, Kim, Mrksic, hao Su, Vandyke, and
  Young}]{Wen2015StochasticLG}
Wen, T.-H.; Gasic, M.; Kim, D.; Mrksic, N.; hao Su, P.; Vandyke, D.; and Young,
  S.~J. 2015.
\newblock Stochastic Language Generation in Dialogue using Recurrent Neural
  Networks with Convolutional Sentence Reranking.
\newblock In \emph{SIGDIAL Conference}.

\bibitem[{Wen et~al.(2016)Wen, Vandyke, Mrksic, Gasic, Rojas-Barahona, Su,
  Ultes, and Young}]{wen2016network}
Wen, T.-H.; Vandyke, D.; Mrksic, N.; Gasic, M.; Rojas-Barahona, L.~M.; Su,
  P.-H.; Ultes, S.; and Young, S. 2016.
\newblock A network-based end-to-end trainable task-oriented dialogue system.
\newblock \emph{arXiv preprint arXiv:1604.04562} .

\bibitem[{Williams, Asadi, and Zweig(2017)}]{Williams2017HybridCN}
Williams, J.~D.; Asadi, K.; and Zweig, G. 2017.
\newblock Hybrid Code Networks: practical and efficient end-to-end dialog
  control with supervised and reinforcement learning.
\newblock \emph{ArXiv} abs/1702.03274.

\bibitem[{Wu et~al.(2018)Wu, Tian, Xia, Fan, Qin, Lai, and
  Liu}]{Wu2018LearningTT}
Wu, L.; Tian, F.; Xia, Y.; Fan, Y.; Qin, T.; Lai, J.-H.; and Liu, T.-Y. 2018.
\newblock Learning to Teach with Dynamic Loss Functions.
\newblock In \emph{NeurIPS}.

\bibitem[{Young et~al.(2010)Young, Gasic, Keizer, Mairesse, Schatzmann,
  Thomson, and Yu}]{Young2010TheHI}
Young, S.~J.; Gasic, M.; Keizer, S.; Mairesse, F.; Schatzmann, J.; Thomson, B.;
  and Yu, K. 2010.
\newblock The Hidden Information State model: A practical framework for
  POMDP-based spoken dialogue management.
\newblock \emph{Computer Speech $\&$ Language} 24: 150--174.

\bibitem[{Zhang, Ou, and Yu(2019)}]{zhang2019task}
Zhang, Y.; Ou, Z.; and Yu, Z. 2019.
\newblock Task-Oriented Dialog Systems that Consider Multiple Appropriate
  Responses under the Same Context.
\newblock \emph{arXiv preprint arXiv:1911.10484} .

\bibitem[{Zhao and Eskenazi(2018)}]{zhao2018zero}
Zhao, T.; and Eskenazi, M. 2018.
\newblock Zero-shot dialog generation with cross-domain latent actions.
\newblock \emph{arXiv preprint arXiv:1805.04803} .

\bibitem[{Zhu(2015)}]{Zhu2015MachineTA}
Zhu, X. 2015.
\newblock Machine Teaching: An Inverse Problem to Machine Learning and an
  Approach Toward Optimal Education.
\newblock In \emph{AAAI}.

\end{thebibliography}
\end{document}